\documentclass{article} 
\usepackage{iclr2027_conference,times}

\usepackage{amsmath,amsfonts,bm}

\def\eqref#1{equation~\ref{#1}}

\def\1{\bm{1}}

\DeclareMathAlphabet{\mathsfit}{\encodingdefault}{\sfdefault}{m}{sl}
\SetMathAlphabet{\mathsfit}{bold}{\encodingdefault}{\sfdefault}{bx}{n}

\usepackage{amssymb}
\usepackage{booktabs}
\usepackage{graphicx}
\usepackage{hyperref}
\usepackage{microtype}
\usepackage{url}

\graphicspath{{figures/}}

\title{
LatentQuant: Preserving the Policy-Facing Latent Contract
under NVFP4 VAE Quantization
}

\newcommand{\affmark}[1]{%
  \textsuperscript{\normalfont\mdseries #1}%
}

\author{
\makebox[\dimexpr\textwidth-2\tabcolsep\relax][c]{%
  \fontsize{11}{13}\selectfont
  \textbf{Ziye Deng}\affmark{1,5}
  \hfill
  \textbf{Lufang Chen}\affmark{5}
  \hfill
  \textbf{Shuyu Feng}\affmark{2}
  \hfill
  \textbf{Zhenwei Duan}\affmark{3,5}
  \hfill
  \textbf{Zicong Ye}\affmark{4,5}
}
\\[5pt]
\makebox[\dimexpr\textwidth-2\tabcolsep\relax][c]{%
  \fontsize{11}{13}\selectfont
  \textbf{Yu Sun}\affmark{2,5}
  \hfill
  \textbf{Xiaofan Li}\affmark{5}
  \hfill
  \textbf{Ruyi Gan}\affmark{5}
  \hfill
  \textbf{Hao Wang}\affmark{5}
  \hfill
  \textbf{Hao Zhang}\affmark{1}\thanks{Corresponding author}
}
\\[11pt]
\makebox[\dimexpr\textwidth-2\tabcolsep\relax][c]{%
  \normalfont\mdseries
  \fontsize{9.5}{11.5}\selectfont
  \affmark{1}UC San Diego
  \qquad
  \affmark{2}Sun Yat-Sen University
  \qquad
  \affmark{3}University of Chinese Academy of Sciences
}
\\[4pt]
\makebox[\dimexpr\textwidth-2\tabcolsep\relax][c]{%
  \normalfont\mdseries
  \fontsize{9.5}{11.5}\selectfont
  \affmark{4}HKUST (Guangzhou)
  \qquad\qquad
  \affmark{5}XSquare Robot
}
}

\iclrfinalcopy

\begin{document}

\maketitle
\lhead{}
\renewcommand{\thefootnote}{\fnsymbol{footnote}}
\renewcommand{\thefootnote}{\arabic{footnote}}

\begin{abstract}
Recent world action models (WAMs) reuse pretrained video VAEs whose encoder
latents directly condition downstream action policies. Quantization must
therefore preserve not only reconstruction fidelity but also the
\textbf{policy-facing latent contract} expected by the frozen policy. Direct
NVFP4 leaves W4A4 quantization error uncompensated, whereas joint
quantization-aware training (QAT) can recover reconstruction by moving this
representation. On Wan2.1, joint QAT nearly matches FP32 VBench-7 (0.7403
versus 0.7409), yet LIBERO success collapses from $95.5\%$ to $10.5\%$.
Controlled decoder-only experiments show that activation quantize--dequantize
operations alter the reconstruction signal and decoder Jacobian, redirecting
the gradient returned to the encoder and inducing persistent latent drift. Based on this mechanism, we introduce \textbf{LatentQuant}, a two-stage NVFP4 QAT framework that first aligns the quantized encoder with its high-precision counterpart, then
freezes it while adapting the decoder. Across Wan2.1 and Wan2.2, LatentQuant
preserves near-baseline control and high reconstruction quality, achieving
$95.75\%$ success on LIBERO and $68.8\%$ on RoboTwin. On NVIDIA B300 GPUs, NVFP4 execution achieves 1.17×–1.26× end-to-end VAE speedups over BF16 cuDNN

\end{abstract}

\section{Introduction}
\label{sec:introduction}

Variational autoencoders (VAEs) are conventionally evaluated by reconstruction fidelity~\citep{yu2023magvitv2,chen2024dcae,nvidia2025cosmostokenizer}. In recent world action models (WAMs), however, the encoder representation can also be consumed directly by a downstream action policy~\citep{cheang2024gr2,li2025uva,li2026lingbotva,ye2026dreamzero}. Closed-loop control further imposes a tight per-step latency budget, making the VAE worth optimizing even when its cost is secondary in offline video generation~\citep{akbari2026actquant,xu2026daptq,zhang2026quantvla}. The latent is therefore both an internal code and a policy-facing representation whose semantics must remain compatible with a pretrained policy, a requirement not captured by reconstruction-centric quantization.

We call this requirement the \textbf{policy-facing latent contract}. Reconstruction constrains the composition of an encoder $E$ and decoder $D$, but not their intermediate coordinates. For any invertible transformation $T$,
\begin{equation}
E'(x)=T(E(x)), \qquad D'(z)=D(T^{-1}(z)),
\end{equation}
the transformed pair preserves reconstruction, whereas a fixed policy $P$ is generally not invariant to the same reparameterization:
\begin{equation}
P(E'(x)) \neq P(E(x)).
\end{equation}
Hence, preserving $D(E(x))$ does not guarantee preservation of the representation consumed by $P$; policy-facing VAE quantization must preserve both reconstruction fidelity and the latent contract.

Standard joint quantization-aware training (QAT)~\citep{jacob2018quantization,huang2025qvgen,zhang2026attnqat} can strongly violate this contract. Under W4A4 NVFP4 QAT~\citep{nvidia2025nvfp4}, the encoder and decoder co-adapt to quantization error by moving the latent while maintaining reconstruction. On Wan2.1~\citep{wanteam2025wan}, joint QAT nearly matches FP32 VBench-7~\citep{huang2024vbench} ($0.7403$ versus $0.7409$), yet LIBERO~\citep{liu2023libero} success falls from $95.5\%$ to $10.5\%$; latent regularization recovers only $86.8\%$. Reconstruction quality can thus conceal severe incompatibility with the frozen policy.

Controlled interventions trace this mismatch to decoder activation quantization. By changing both the reconstruction signal and the decoder Jacobian, quantization redirects the gradient returned to the encoder; repeated optimizer updates consolidate this signal into a persistent latent shift. The failure is therefore a learned drift of the policy-facing representation, not merely an instantaneous perturbation at the encoder output.

\begin{figure*}[t]
    \centering
    \includegraphics[width=\textwidth,trim=0 4 0 3,clip]{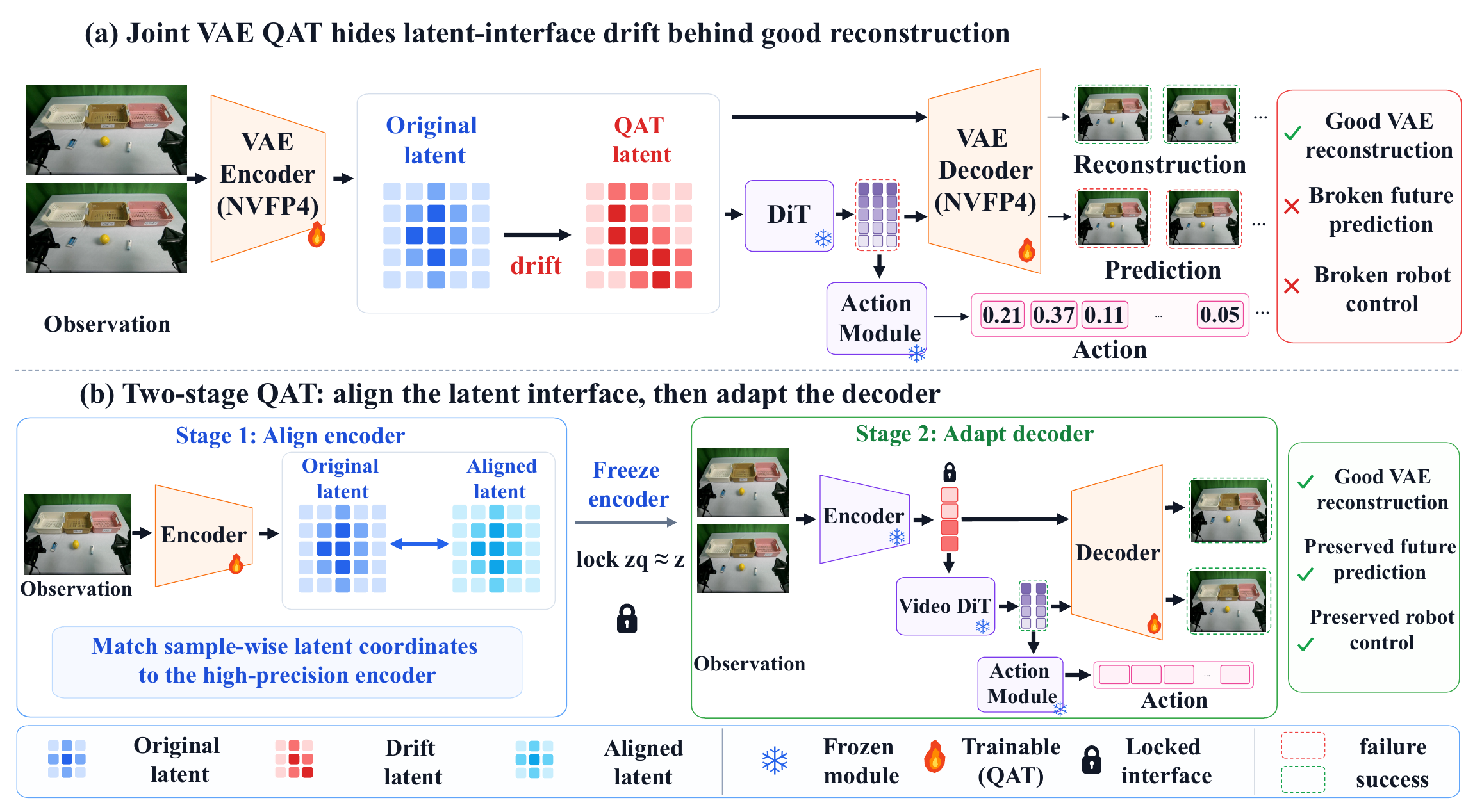}
    \caption{LatentQuant separates encoder alignment from decoder adaptation.
    Stage~I aligns the NVFP4 encoder with a frozen high-precision reference; Stage~II
    freezes this interface and adapts only the NVFP4 decoder, preventing
    reconstruction gradients from rewriting the policy-facing latent.}
    \label{fig:overview}
\end{figure*}

This analysis suggests a hard design constraint: reconstruction adaptation should not rewrite an established latent contract. We therefore propose \textbf{LatentQuant}, the two-stage NVFP4 QAT framework in Figure~\ref{fig:overview}. Stage~I aligns the quantized encoder with its high-precision reference under NVFP4 execution. Stage~II freezes the aligned encoder and adapts only the quantized decoder for reconstruction. This boundary permits low-precision adaptation while preventing decoder reconstruction gradients from modifying the policy-facing latent.

Across Wan2.1 and Wan2.2~\citep{wanteam2025wan}, LatentQuant retains near-baseline control and high reconstruction quality: $95.75\%$ versus $95.50\%$ success on LIBERO, and $68.8\%$ versus $69.8\%$ on RoboTwin~\citep{chen2025robotwin}. A native packed-NVFP4 convolution backend on NVIDIA B300 GPUs further demonstrates that contract preservation is compatible with end-to-end VAE acceleration.

Our contributions are summarized as follows:
\begin{itemize}
    \item We formulate the \textbf{policy-facing latent contract} for quantizing VAEs consumed by fixed downstream policies, and show that reconstruction fidelity can conceal severe policy incompatibility.
    \item We identify an \textbf{optimization-induced latent-contract violation mechanism} under joint low-bit QAT, showing that decoder-side quantization redirects reconstruction gradients returned to the encoder and consolidates them into persistent learned latent drift.
    \item We propose \textbf{LatentQuant}, a two-stage QAT method which separates quantized encoder alignment from decoder adaptation to preserve the established latent contract. On NVIDIA B300 GPUs, NVFP4 execution achieves \textbf{1.17$\times$--1.26$\times$} end-to-end VAE speedups over BF16 cuDNN.
\end{itemize}

\section{Related Work} \label{sec:related-work} \textbf{WAMs and policy-consuming latents.} Recent video-action models couple latent visual representations with action prediction through diverse architectures. A growing class of WAMs builds directly on latent video generators and inherits their pretrained VAEs. GR-2 jointly predicts future videos and actions, while UVA learns a joint video--action latent representation with separate decoding heads~\citep{cheang2024gr2,li2025uva}. More recent WAMs, including WALL-WM, LingBot-VA, and DreamZero, further couple video generation and action prediction through different unified or causal architectures~\citep{li2026wallwm,li2026lingbotva,ye2026dreamzero}. Despite the various WAM architectures, the latents produced from the encoder of VAEs are consumed by the downstream action policy. As a result, it is critical to preserve the latent while quantizing VAEs. This compatibility concern also appears in video-VAE design, where latent-space alignment is used to retain compatibility with a pretrained generator~\citep{zhao2024cvvae,cai2026davae}. Our setting is stricter: the downstream action policy is frozen, so the established latent coordinates cannot be adapted after quantization.

\textbf{Low-bit quantization for generative models.}
Existing quantization studies for generative models have primarily focused on the Transformer backbone. Q-Diffusion, SVDQuant, and ViDiT-Q develop low-bit quantization methods for diffusion models, while QVGen introduces QAT for video diffusion models at 4-bit and below~\citep{li2023qdiffusion,zhao2025viditq,li2025svdquant,huang2025qvgen}. In contrast, VAE quantization has received substantially less attention than the Transformer backbone in low-bit diffusion pipelines. PassionSR~\citep{zhu2025passionsr} is among the few works that explicitly quantize the VAE, but it focuses on W8A8/W6A6 PTQ for image super-resolution. Complementary video-VAE work targets compression and decoder efficiency rather than low-bit weight-and-activation execution~\citep{wu2025h3ae,zhu2026flashvaed}. Although PTQ is time efficient, QAT can achieve higher accuracy, which is critical for downstream action policy. With native NVFP4 support in NVIDIA's low-precision training stack~\citep{nvidia2025nvfp4}, an effective VAE-specific NVFP4 QAT method is needed for practical low-bit acceleration.

\textbf{Interface-aware quantization.}
Recent work has shown that low-bit quantization can affect intermediate representations and downstream behavior, not only final outputs. Quantized feature distillation and quantization-aware distillation use teacher-guided objectives to mitigate such degradation~\citep{zhu2023qfd,xin2026qad}, while QuantWM, QuantVLA, ActQuant, and DA-PTQ study quantization-induced degradation in world models and vision-language-action policies~\citep{fu2026worldquant,zhang2026quantvla,akbari2026actquant,xu2026daptq}. Analyses of nonuniform floating-point training also identify systematic shrinkage bias in low-bit optimization~\citep{zhao2026ufp4}. Studies of latent geometry and frequency structure further show that a reconstruction-preserving latent can still be unfavorable to a downstream generator~\citep{yao2025vavae,chen2025dcae15,skorokhodov2025diffusability,kouzelis2025eqvae}. In contrast, we focus on a different failure mode arising during joint QAT of a pretrained video VAE, where reconstruction optimization persistently shifts the policy-facing encoder representation. We therefore formulate latent preservation as a representation contract during low-bit VAE adaptation.

\section{Problem Setting}
\label{sec:problem_setting}

\subsection{Policy-Facing VAE Quantization}
\label{sec:policy_facing_vae}

We consider WAMs in which the representation produced by a pretrained video VAE
conditions a fixed downstream action policy. Given an observation clip $x$, the
high-precision encoder $E_0$ produces
\begin{equation}
    z_0 = E_0(x),
\end{equation}
and the corresponding action prediction is abstracted as
\begin{equation}
    a_0 = P(z_0, c),
\end{equation}
where $c$ denotes the task condition and $P$ remains frozen throughout VAE
quantization.

Let $E_\theta^q$ and $D_\phi^q$ denote the quantized encoder and decoder, with
trainable parameters $\theta$ and $\phi$, and let
\begin{equation}
    z_q = E_\theta^q(x)
\end{equation}
denote the quantized encoder representation. Conventional VAE quantization
primarily seeks to preserve reconstruction through
$D_\phi^q(E_\theta^q(x))$. In the policy-facing setting, this objective is
insufficient: the quantized representation must also remain compatible with
the representation expected by the fixed downstream policy. We therefore view
VAE quantization as reconstruction adaptation subject to preservation of the
policy-facing latent contract,
\begin{equation}
    E_\theta^q(x) \simeq E_0(x),
\end{equation}
where the notion of latent deviation used for evaluation is specified in
Section~\ref{sec:experiments}.

\subsection{NVFP4 Quantization}
\label{sec:nvfp4_setting}

We study W4A4 quantization under the Blackwell NVFP4 format
\citep{nvidia2025nvfp4}. NVFP4 represents data values in E2M1 and uses
two-level scaling consisting of an FP32 per-tensor scale and an E4M3 scale for
each contiguous 16-element block. We apply NVFP4 QDQ to the weight and
activation operands of convolutional and linear layers. Normalization,
nonlinearities, residual additions, and the VAE input and latent boundaries
remain in higher precision.

During QAT, master parameters and optimizer states remain in FP32, while the
selected operands follow the NVFP4 QDQ contract in the forward pass with a
straight-through estimator for backpropagation. Direct quantization uses the
same arithmetic and layer coverage without parameter adaptation. All low-bit
baselines therefore share the same numerical contract, isolating the effect of
the optimization strategy. Detailed scaling rules, operand layouts, and
mixed-precision scopes are provided in Appendix~\ref{app:nvfp4-details}.

\section{Mechanism of Latent-Contract Violation under Joint QAT}
\label{sec:mechanism}

\begin{figure*}[t]
    \centering
    \includegraphics[width=\textwidth]{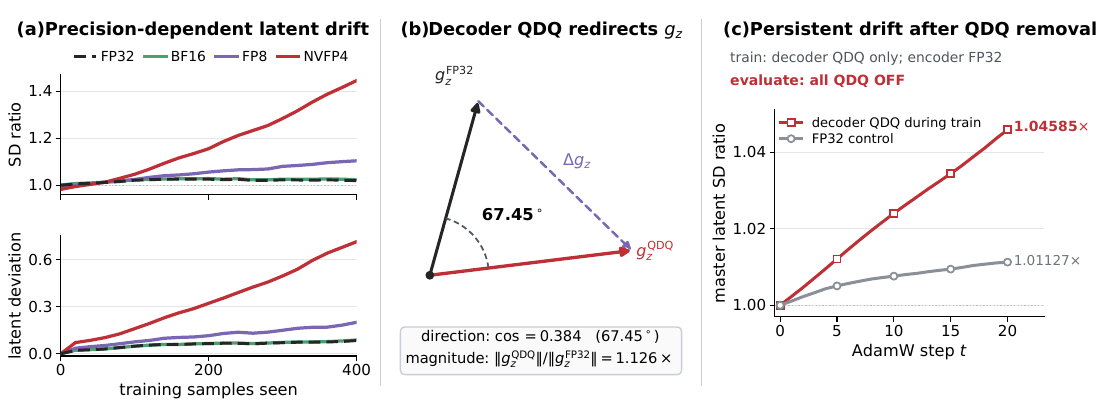}
    \caption{Mechanism of latent-contract violation under joint QAT.
    Low precision progressively shifts the learned latent (a); decoder activation
    QDQ redirects its reconstruction gradient (b); and the drift persists after
    removing QDQ, showing consolidation in encoder parameters (c).}
    \label{fig:mechanism}
\end{figure*}

Joint QAT optimizes the encoder and decoder jointly under reconstruction objectives.
Although this flexibility compensates for quantization errors, it also allows
reconstruction adaptation to modify the policy-facing latent representation. We
analyze this optimization pathway through three stages: latent drift under low
precision, decoder-induced gradient perturbation, and persistent encoder
representation change.

\subsection{Precision-Dependent Latent Drift}
\label{sec:precision_drift}

We first characterize how low-precision QAT changes the learned latent
representation. Given encoder representation $z_t(x)$ at training step $t$, we
measure the relative latent-scale variation and sample-wise deviation as

\begin{equation}
r_{\mathrm{SD}}(t)
=
\frac{
\operatorname{Std}[z_t(x)]
}{
\operatorname{Std}[z_0(x)]
},
\qquad
d_{\mathrm{rel}}(t)
=
\mathbb{E}_{x}
\left[
\frac{
\|z_t(x)-z_0(x)\|_2
}{
\|z_0(x)\|_2
}
\right]
\label{eq:latent_drift_metrics}
\end{equation}

where $z_0(x)$ denotes the original latent representation before QAT. Under
matched training trajectories, FP32 and BF16 remain close to the original
representation, while FP8 and NVFP4 introduce progressively larger deviations.
As shown in Figure~\ref{fig:mechanism}(a), after 400 matched samples, NVFP4 reaches
an SD ratio of $1.445\times$ and a sample-wise deviation of $0.714$, compared with
$1.020\times$ and $0.084$ in FP32.

NVFP4 QDQ additionally introduces systematic magnitude contraction in decoder
activations across most measured sites, consistent with the shrinkage-bias
phenomenon reported for nonuniform FP4 grids~\citep{zhao2026ufp4}. We treat this as
a numerical property of the structured quantization perturbation rather than
attributing latent drift to a single scalar shrinkage effect. Since QDQ perturbations depend on quantization bins,
block scales, and activation distributions, we next analyze how decoder-side
quantization modifies the optimization signal propagated to the encoder.

\subsection{Decoder Quantization Redirects Latent Gradients}
\label{sec:gradient_redirection}

Decoder activation quantization changes the reconstruction pathway before the
gradient reaches the encoder. We denote the latent gradients under full-precision
and quantized decoder execution as

\begin{equation}
g_z^0
=
J_0^\top u_0,
\qquad
g_z^q
=
J_q^\top u_q
\label{eq:latent_gradient}
\end{equation}

where $u$ denotes the reconstruction-loss gradient at the decoder output and
$J^\top$ denotes the corresponding decoder vector-Jacobian product.

The latent-gradient difference can be decomposed as

\begin{equation}
\Delta g_z
=
g_z^q-g_z^0
=
\underbrace{
J_0^\top
(u_q-u_0)
}_{\text{reconstruction-signal change}}
+
\underbrace{
(J_q^\top-J_0^\top)u_q
}_{\text{backward-map change}}
\label{eq:gradient_decomposition}
\end{equation}

The first term corresponds to the change in reconstruction correction, while the
second captures the altered backward transport induced by the quantized decoder.

A controlled decoder-side intervention validates this mechanism. Applying NVFP4 QDQ
only to decoder activations changes the reconstruction objective, increasing the
matched reconstruction loss from $0.008834$ to $0.017591$. Further restricting QDQ
to the activation before the first decoder convolution produces an $11.857\%$
difference in the final reconstruction. As shown in Figure~\ref{fig:mechanism}(b),
the resulting latent gradient rotates by $67.45^\circ$ despite only a $1.126\times$
increase in gradient norm. Therefore, decoder quantization primarily redirects the
optimization direction rather than simply amplifying the gradient magnitude.

The complete layerwise sweep, decomposition closure, and reverse-reference analysis
are provided in Appendix~\ref{app:mechanism-details}.

\subsection{Gradient Perturbation Leads to Persistent Latent Drift}
\label{sec:persistent_drift}

The altered latent gradient affects the learned representation only if it propagates
through the encoder parameterization. Since the compared branches share the same
high-precision encoder, the corresponding encoder parameter gradients satisfy

\begin{equation}
g_{\theta}^{0}
=
J_E^\top g_z^0,
\qquad
g_{\theta}^{q}
=
J_E^\top g_z^q
\label{eq:encoder_gradient}
\end{equation}

where $J_E^\top$ denotes the encoder vector-Jacobian product. The latent-gradient
perturbation is therefore transferred to encoder parameter updates. In the controlled
intervention, the $1.126\times$ latent-gradient norm ratio becomes a $1.514\times$
encoder-gradient norm ratio, confirming that the decoder-induced perturbation reaches
the trainable encoder parameters.

To distinguish learned drift from instantaneous quantization effects, we evaluate the
high-precision master encoder after removing all QDQ operators:

\begin{equation}
z_t^{M}
=
E_{\theta_t}^{0}(x_{\mathrm{probe}})
\label{eq:master_latent}
\end{equation}

As shown in Figure~\ref{fig:mechanism}(c), although encoder quantization is never
enabled, the decoder-QDQ branch produces a master-latent SD ratio of $1.04585$,
compared with $1.01127$ in the high-precision branch. Since the evaluation removes
all QDQ operators, this additional deviation reflects a learned change in encoder
parameters rather than an instantaneous quantization artifact.

Together, these controlled interventions establish the following sufficient pathway:

\begin{equation}
\mathrm{QDQ}_{D}
\rightarrow
\Delta g_z
\rightarrow
\Delta g_{\theta}
\rightarrow
\Delta\theta
\rightarrow
\Delta z^{M}
\label{eq:causal_path}
\end{equation}

Decoder-side quantization can therefore violate the latent contract because
reconstruction optimization is allowed to rewrite the encoder representation consumed
by the frozen downstream policy.

\section{From Mechanism to Method: LatentQuant}
\label{sec:method}

The analysis in Section~\ref{sec:mechanism} yields a direct design principle: reconstruction adaptation must not be allowed to rewrite an established policy-facing latent contract. An alternative is to optimize a joint reconstruction and sample-wise alignment objective:

\begin{equation}
\min_{\theta,\phi}
\;
\mathbb{E}_{x\sim\mathcal{D}}
\left[
\mathcal{L}_{\mathrm{rec}}
\left(
D_\phi^q(E_\theta^q(x)),x
\right)
+
\lambda
\mathcal{L}_{\mathrm{align}}
\left(
E_\theta^q(x),E_0(x)
\right)
\right]
\label{eq:soft_joint_objective}
\end{equation}

Such a joint objective can suppress latent drift, but leaves encoder coordinates subject to both reconstruction and alignment updates. LatentQuant instead first establishes an aligned low-bit encoder and then freezes it during decoder adaptation. This explicitly blocks further reconstruction-driven encoder updates, rather than relying on the joint objective to maintain the established interface. Stage~I aligns the NVFP4 encoder with the high-precision
reference; Stage~II freezes the aligned encoder and restricts reconstruction
adaptation to the decoder.

With $z_0=\operatorname{sg}(E_0(x))$ and
$z_q(\theta)=E_\theta^q(x)$, the two stages are

\begin{equation}
\theta^\star
=
\arg\min_{\theta}
\;
\mathbb{E}_{x\sim\mathcal{D}}
\left[
\mathcal{L}_{\mathrm{align}}
\left(
z_q(\theta),z_0
\right)
\right]
\label{eq:stage1_objective}
\end{equation}

\begin{equation}
\phi^\star
=
\arg\min_{\phi}
\;
\mathbb{E}_{x\sim\mathcal{D}}
\left[
\mathcal{L}_{\mathrm{rec}}
\left(
D_\phi^q
\left(
\operatorname{sg}(z_q(\theta^\star))
\right),
x
\right)
\right]
\quad
\mathrm{s.t.}\quad
\theta=\theta^\star
\label{eq:stage2_objective}
\end{equation}

The Stage-II constraint is the key distinction: latent preservation is no longer a
soft objective traded against reconstruction, but an explicit optimization boundary.

\subsection{Stage I: Sample-Wise Encoder Alignment}
\label{sec:stage1}

Stage~I adapts the encoder to NVFP4 execution while anchoring its output to the
reference representation expected by the frozen downstream policy. The reference
encoder remains fixed, while NVFP4 QDQ is active throughout the trainable encoder
path.

Because latent channels have different natural scales, we normalize the alignment
error by the reference channel-wise standard deviation. Let $\sigma_c$ denote the
standard deviation of latent channel $c$ over the training distribution and
$\tilde{\sigma}_c=\max(\sigma_c,\epsilon)$. For a latent tensor of shape
$B\times C\times T_z\times H_z\times W_z$, we minimize

\begin{equation}
\mathcal{L}_{\mathrm{align}}
=
\frac{1}{2BCT_zH_zW_z}
\sum_{b,c,\tau,h,w}
\left(
\frac{
z_{q,bc\tau hw}
-
z_{0,bc\tau hw}
}{
\tilde{\sigma}_c
}
\right)^2
\label{eq:alignment}
\end{equation}

Channel normalization prevents high-variance dimensions from dominating the
objective. The alignment is also sample-wise rather than distribution-level:
matching aggregate statistics can preserve marginal distributions while allowing
content-dependent representation shifts, whereas Equation~\ref{eq:alignment}
anchors each observation to its corresponding pretrained latent coordinates.

This coordinate-level alignment is an operational constraint for preserving the
latent contract rather than a claim that exact coordinate matching is the only form
of policy compatibility. Since $E_0$ already produces representations compatible
with the frozen policy, it provides a direct supervision target without requiring
action labels or policy gradients. Only encoder parameters are updated in Stage~I;
the decoder is not involved in the optimization.

\subsection{Stage II: Decoder Adaptation under a Fixed Contract}
\label{sec:stage2}

Stage~II retains the aligned encoder under the same NVFP4 execution used at
deployment but freezes its master parameters. The decoder therefore observes the
actual quantized latent distribution, while reconstruction error must be absorbed
entirely on the decoder side of the latent boundary.

We optimize a composite reconstruction objective

\begin{equation}
\mathcal{L}_{\mathrm{rec}}
=
\mathcal{L}_{\mathrm{MSE}}
+
\lambda_{\mathrm{edge}}
\mathcal{L}_{\mathrm{Sobel}}
+
\lambda_{\mathrm{perc}}
\mathcal{L}_{\mathrm{LPIPS}}
\label{eq:reconstruction_objective}
\end{equation}

with loss weights, optimization schedules, and checkpoint-selection details reported
in Appendix~\ref{app:training-evaluation}. Because $\theta=\theta^\star$ throughout this stage, decoder
reconstruction gradients have no path through which to introduce additional learned
encoder drift.

Stage~I therefore compensates for encoder-side NVFP4 error while preserving the
reference representation, and Stage~II recovers reconstruction quality on the fixed
latent contract. Both VAE components adapt to low-precision arithmetic, but
reconstruction optimization is prevented from renegotiating the representation
consumed by the frozen downstream policy.

\section{Experiments}
\label{sec:experiments}

We evaluate LatentQuant on whether reconstruction quality predicts policy compatibility, whether separating encoder alignment from decoder adaptation recovers reconstruction while retaining control, and whether the resulting VAE supports native NVFP4 execution.

\subsection{Experimental Setup}
\label{sec:exp_setup}

We evaluate two WALL-WM~\citep{li2026wallwm} variants that consume pretrained
video-VAE latents for control. Following their native configurations, Wan2.1 is
evaluated on LIBERO and Wan2.2 on RoboTwin, with FP32 and BF16 references,
respectively. We additionally perform paired replay on fixed physical
observations collected with the XSquare Robot platform.

We compare \textbf{FP32/BF16}, \textbf{Direct NVFP4} without adaptation,
\textbf{Joint QAT}, \textbf{Joint+KL}, and \textbf{LatentQuant}. All low-bit
variants share the same NVFP4 W4A4 arithmetic and mixed-precision scope, while
the downstream policies remain frozen; their differences therefore isolate
the optimization strategy.
Joint+KL augments joint QAT with $0.01\,D_{\mathrm{KL}}[\mathcal{N}(\mu_q,\sigma_q^2)\,\|\,\mathcal{N}(\mu_{\mathrm{ref}},\sigma_{\mathrm{ref}}^2)]$, averaged over latent channels, where the moments are computed over the batch and spatiotemporal axes; unlike Stage~I, it does not align corresponding samples.
Following VBench~\citep{huang2024vbench}, we report VBench-7, the unweighted
mean of seven VBench dimensions, for
reconstruction, channel-normalized latent deviation $d_z$, standardized action
deviation $d_a$, and closed-loop success. The paired $d_z$/$d_a$ evaluation
holds observations, task conditions, action noise, and policy state fixed.
Success is measured over 800 LIBERO and 500 RoboTwin trials. Full protocols are
provided in Appendices~\ref{app:training-evaluation}--\ref{app:downstream-details}.

\subsection{Reconstruction Does Not Determine Policy Compatibility}
\label{sec:main_results}

\begin{table*}[t]
\centering
\caption{
Main comparison across Wan2.1 and Wan2.2. VBench-7 measures reconstruction
quality, while $d_z$ and $d_a$ measure latent and action deviations.
Lower $d_z$ and $d_a$ are better.
}
\label{tab:main-results}
\small
\begin{tabular}{llccccc}
\toprule
VAE & Method &
Robot VBench-7 $\uparrow$ &
LIBERO VBench-7 $\uparrow$ &
$d_z \downarrow$ &
$d_a \downarrow$ &
Success $\uparrow$ \\
\midrule

Wan2.1
& FP32
&0.7409&0.7747&0.000&0.000&95.50\\
& Direct NVFP4
&0.7266&0.7513&0.288&0.033&94.38\\
& Joint QAT
&0.7403&0.7565&4.796&0.270&10.50\\
& Joint+KL
&0.7341&0.7571&1.190&0.118&86.75\\
& LatentQuant
&0.7340&0.7513&0.202&0.018&95.75\\

\midrule

Wan2.2
& BF16
&0.7436&0.7750&0.000&0.000&69.80\\
& Direct NVFP4
&0.7325&0.7626&0.092&0.202&68.00\\
& Joint QAT
&0.7402&0.7613&4.638&0.344&0\\
& Joint+KL
&0.7407&0.7638&0.995&0.340&8.40\\
& LatentQuant
&0.7414&0.7656&0.062&0.003&68.80\\

\bottomrule
\end{tabular}
\end{table*}

Table~\ref{tab:main-results} reveals the same reconstruction--control mismatch
in both WAM stacks. On Wan2.1, Joint QAT nearly matches the high-precision
Robot VBench-7 score ($0.7403$ versus $0.7409$), yet LIBERO success collapses
from $95.50\%$ to $10.50\%$. On Wan2.2, it improves Robot VBench-7 over
Direct NVFP4 ($0.7402$ versus $0.7325$) while reducing RoboTwin success from
$69.80\%$ to $0\%$. Conversely, Direct NVFP4 preserves $94.38\%$ and
$68.00\%$ success despite larger reconstruction degradation. Reconstruction-only
selection would therefore rank policy-incompatible adapted VAEs above the
control-preserving Direct baseline.

The paired metrics expose this failure before rollouts diverge. Joint QAT
produces $d_z=4.796/4.638$ and $d_a=0.270/0.344$ on Wan2.1/Wan2.2,
whereas LatentQuant reduces them to $0.202/0.062$ and $0.018/0.003$,
respectively, smaller than Direct NVFP4 on both stacks. LatentQuant consequently
retains near-reference control: $95.75\%$ versus $95.50\%$ on LIBERO and
$68.80\%$ versus $69.80\%$ on RoboTwin. It also improves Robot VBench-7 over
Direct NVFP4 from $0.7266$ to $0.7340$ and from $0.7325$ to $0.7414$. The two
additional LIBERO successes indicate parity rather than an improvement from
quantization.

\subsection{Why the Optimization Boundary Matters}
\label{sec:ablation}

\begin{table}[t]
    \centering
    \caption{\textbf{Stage-I and boundary ablation on Wan2.1.}
    Freeze-Only omits Stage-I alignment and therefore retains the Direct NVFP4 interface metrics.}
    \label{tab:stage1-ablation}
    \small
    \setlength{\tabcolsep}{8pt}
    \renewcommand{\arraystretch}{1.02}
    \begin{tabular*}{\textwidth}{@{\extracolsep{\fill}}lccc@{}}
        \toprule
        Metric & \textsc{Direct} & \textsc{Freeze-Only} & \textsc{LatentQuant} \\
        \midrule
        Robot VBench-7 $\uparrow$  & .7266 & .7326 & .7340 \\
        LIBERO VBench-7 $\uparrow$ & .7513 & .7512 & .7513 \\
        $d_z \downarrow$           & .288  & .288  & .202  \\
        $d_a \downarrow$           & .033  & .033  & .018  \\
        \bottomrule
    \end{tabular*}
\end{table}

Table~\ref{tab:stage1-ablation} separates the two roles of LatentQuant.
Freeze-Only adapts only the decoder, improving reconstruction while retaining
the Direct NVFP4 interface deviation because its encoder is unchanged.
LatentQuant first aligns the encoder and then freezes it, so Stage~I removes
the initial mismatch and the Stage-II boundary prevents reconstruction updates
from reintroducing latent drift.

The matched soft-alignment sweep in Appendix~\ref{app:lambda-sweep}
(Table~\ref{tab:appendix-lambda-sweep} and Figure~\ref{fig:lambda-sweep})
shows the same trade-off directly: $\lambda=0$ yields
$(d_z,d_a)=(1.115,0.071)$, while any $\lambda\geq0.03$ reduces both to at
most $(0.222,0.040)$. Increasing the weight beyond $\lambda=0.3$ provides
little additional interface improvement but lowers held-out PSNR from
$29.77$ to $29.19$ dB, motivating a separate alignment stage rather than a
single finite joint-QAT penalty.

\subsection{Physical-Observation Transfer and Native Deployment}
\label{sec:deployment}

\begin{figure*}[t]
\centering
\includegraphics[width=0.98\textwidth]{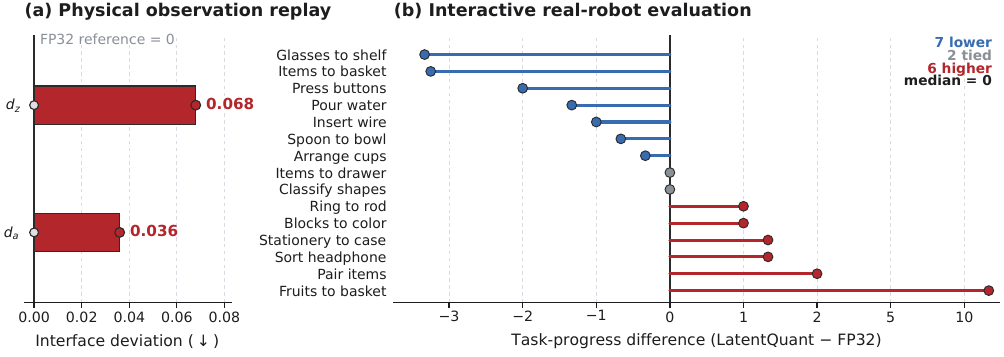}
\caption{Physical-domain validation. (a) Interface deviations on fixed
observations. (b) LatentQuant--FP32 differences in three-run mean milestone
scores over 15 real-robot tasks: 7 lower, 2 tied, and 6 higher, with median 0.
The positive tail uses a nonuniform scale.}
\label{fig:physical-domain-validation}
\end{figure*}

On fixed physical observations, Figure~\ref{fig:physical-domain-validation}
shows that LatentQuant maintains small interface deviations ($d_z=0.068$ and
$d_a=0.036$). In interactive real-robot evaluation, task-level changes occur
in both directions relative to FP32, with a median task-wise score difference of zero.

\begin{figure}[t]
\centering
\includegraphics[width=\linewidth]{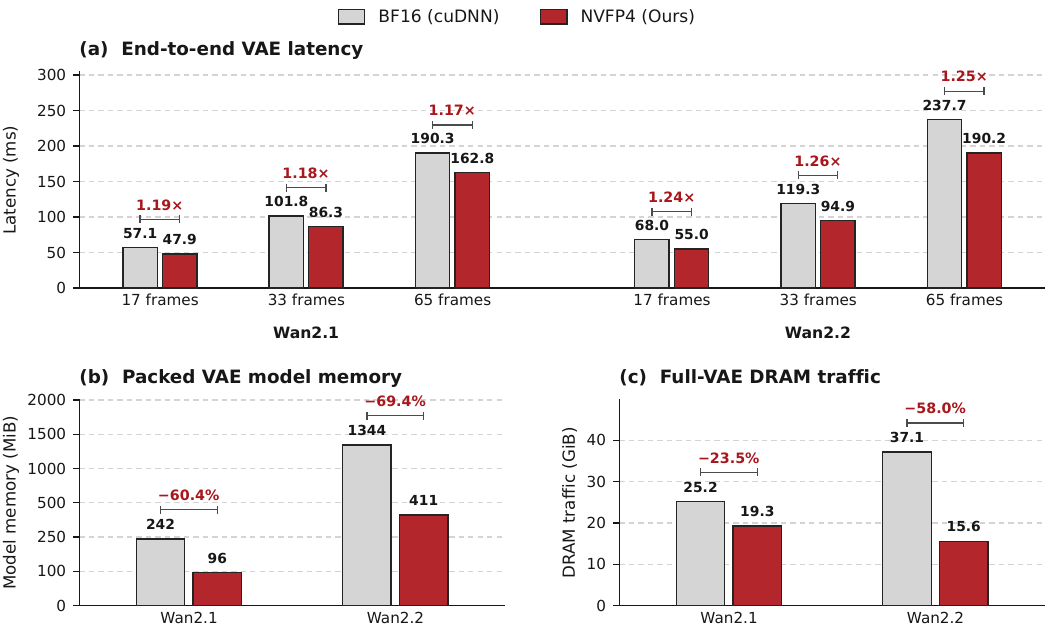}
\caption{
Native NVFP4 VAE deployment efficiency on NVIDIA B300.
}
\label{fig:nvfp4_efficiency}
\end{figure}

Across 17-, 33-, and 65-frame inputs, Figure~\ref{fig:nvfp4_efficiency} shows
$1.17\times$--$1.19\times$ end-to-end speedups for Wan2.1 and
$1.24\times$--$1.26\times$ for Wan2.2 over BF16 cuDNN. Packed model memory is
reduced by $60.4\%$ and $69.4\%$, and full-VAE DRAM traffic by $23.5\%$ and
$58.0\%$, respectively. These measurements include online activation scaling
and packing and all operations that remain in higher precision. Since Direct
NVFP4 and LatentQuant use the same backend, the systems gains come from native
execution; LatentQuant makes them usable without sacrificing the policy-facing
latent contract. Implementation and profiling details are provided in
Appendix~\ref{app:systems-details}.

\section{Conclusion}
\label{sec:conclusion}

Quantizing a WAM VAE is not solely a reconstruction problem: the encoder must
also preserve the representation expected by a frozen action policy. We trace
joint-QAT failure to decoder activation QDQ, which redirects encoder updates and
produces persistent latent drift. LatentQuant prevents further reconstruction-driven encoder drift by aligning the quantized encoder to its high-precision reference, freezing that interface, and then adapting the decoder. Across Wan2.1 and Wan2.2, the resulting NVFP4 VAE
retains visual and closed-loop performance near the high-precision systems,
while a native Blackwell backend reduces latency, memory, and data movement.
Evaluation covers two WAM stacks, LIBERO/RoboTwin, an NVIDIA B300, and
convolutional W4A4; broader models, hardware, and operators remain open.
More broadly, low-bit optimization of shared generative components should
preserve the representations consumed downstream, not only reconstructed outputs.

\section*{AI Use Statement}

Generative AI tools were used to assist with literature search, research and experimental planning, code development, mathematical reasoning, figure preparation, drafting, and language editing. All AI-assisted outputs were reviewed and verified by the authors, including cited literature, mathematical claims, code, and reported results.

\bibliography{iclr2027_conference}
\bibliographystyle{iclr2027_conference}

\clearpage
\appendix
\setcounter{topnumber}{5}
\setcounter{bottomnumber}{4}
\setcounter{totalnumber}{7}
\renewcommand{\topfraction}{0.95}
\renewcommand{\bottomfraction}{0.90}
\renewcommand{\textfraction}{0.05}
\renewcommand{\floatpagefraction}{0.88}
\setlength{\textfloatsep}{10pt plus 2pt minus 2pt}
\setlength{\floatsep}{9pt plus 2pt minus 2pt}
\setlength{\intextsep}{9pt plus 2pt minus 2pt}

\section{NVFP4 Arithmetic and Execution Contract}
\label{app:nvfp4-details}

\subsection{Operand Layout and Mixed-Precision Scope}
\label{app:nvfp4-layout}

The logical NVFP4 block layout follows the convolution dataflow. Activations
are arranged with channels contiguous at each spatiotemporal position, yielding
one E4M3 scale for each $1\times16$ channel block. Convolution weights are
flattened by output channel, with input channels contiguous within each spatial
or temporal kernel tap. Incomplete groups are zero padded. The same rule is
used for Conv2D, Conv3D, and linear weight and activation operands; zero tensors
and underflowed blocks are handled explicitly.

The outgoing latent projection belongs to the encoder parameter scope, whereas
the incoming latent projection belongs to the decoder scope. The two
projections are updated in their respective training stages but remain
higher-precision boundaries in the forward path. Raw video inputs, resizing,
normalization layers, nonlinearities, residual additions, and the latent
interface also remain in higher precision. Throughout the paper, NVFP4 W4A4
therefore denotes this explicit mixed-precision execution contract rather than
indiscriminate FP4 execution of every VAE operation.

\subsection{Training and Deployment State}
\label{app:nvfp4-state}

During QAT, master parameters and optimizer states remain in FP32, while the
selected weight and activation operands are passed through strict NVFP4 QDQ in
the forward graph. Deployment instead stores supported convolution weights in
packed NVFP4 form together with their FP32 tensor scales and E4M3 block scales;
activation scaling and packing are performed online. Direct NVFP4, joint QAT, and LatentQuant use the same arithmetic, operand layout, and layer coverage. Differences in reconstruction and downstream behavior therefore arise from parameter adaptation rather than from different quantizers or mixed-precision scopes.

\section{Optimization and Evaluation Protocols}
\label{app:training-evaluation}

\subsection{Two-Stage Optimization}
\label{app:two-stage-optimization}

All QAT runs start from the corresponding pretrained Wan VAE and use
17-frame, $208\times224$ robot-manipulation clips. The video backbone and
action model remain frozen throughout. Stage~I checkpoints are selected by
validation alignment loss; Stage~II checkpoints are selected on held-out
reconstruction. Table~\ref{tab:appendix-training-config} reports the
optimization settings used for the two-stage models. Batch sizes are per GPU,
and all reported runs use five GPUs.

\begin{table}[!htbp]
    \centering
    \caption{\textbf{Two-stage optimization configuration.}
    Stage~I updates only the encoder using sample-wise latent alignment.
    Stage~II freezes the aligned encoder and updates only the decoder using the
    reconstruction objective.}
    \label{tab:appendix-training-config}
    \small
    \setlength{\tabcolsep}{5pt}
    \renewcommand{\arraystretch}{1.08}
    \begin{tabular*}{\textwidth}{@{\extracolsep{\fill}}llcccl@{}}
        \toprule
        VAE & Trainable block & Epochs & Batch/GPU & LR schedule & Objective \\
        \midrule
        Wan2.1 & Encoder & 1 & 10
            & $5\!\times\!10^{-6}\!\rightarrow\!1\!\times\!10^{-6}$
            & $\mathcal{L}_{\mathrm{align}}$ \\
        Wan2.1 & Decoder & 5 & 16
            & $5\!\times\!10^{-6}\!\rightarrow\!1\!\times\!10^{-6}$
            & $\mathcal{L}_{\mathrm{rec}}$ \\
        Wan2.2 & Encoder & 1 & 10
            & $5\!\times\!10^{-6}\!\rightarrow\!1\!\times\!10^{-6}$
            & $\mathcal{L}_{\mathrm{align}}$ \\
        Wan2.2 & Decoder & 5 & 12
            & $5\!\times\!10^{-6}\!\rightarrow\!1\!\times\!10^{-6}$
            & $\mathcal{L}_{\mathrm{rec}}$ \\
        \bottomrule
    \end{tabular*}
\end{table}

The reconstruction objective is
\[
\mathcal{L}_{\mathrm{rec}}
=
\mathcal{L}_{\mathrm{MSE}}
+
0.05\,\mathcal{L}_{\mathrm{Sobel\text{-}L1}}
+
0.1\,\mathcal{L}_{\mathrm{LPIPS}}.
\]
Both stages use weight decay $10^{-4}$, cosine learning-rate decay,
100 warm-up steps, gradient clipping at $1.0$, and seed 42. LPIPS is
evaluated on four sampled frames per clip. Stage~I does not execute the
decoder and uses neither pixel nor perceptual supervision. Stage~II evaluates
the aligned quantized encoder under stop-gradient and updates decoder master
parameters only.

\subsection{Soft Alignment Sweep}
\label{app:lambda-sweep}

We additionally run a matched diagnostic sweep for a jointly trained encoder
and decoder. The objective is the reconstruction objective plus
$\lambda\mathcal{L}_{\mathrm{align}}$, where
$\lambda\in\{0,0.03,0.1,0.3,1,3,10\}$. All seven runs use the same Wan2.1
initialization, strict NVFP4 W4A4 scope, 64-clip training subset, sample order,
optimizer, schedule, and 1,000 optimization steps. The quantities below are
computed on the same 64 held-out clips at the final step. ``SD ratio'' is the
mean channel-wise latent standard deviation divided by that of the pristine
FP32 encoder; ``max SD'' is the largest channel-wise ratio. ``Moment KL'' is
the Gaussian moment divergence used by the validation diagnostic, and ``mean
gap'' is the mean absolute channel-mean difference. This sweep is intended to
expose the soft-constraint trade-off; it is not an additional downstream
success evaluation.

\begin{table}[!htbp]
    \centering
    \caption{\textbf{Wan2.1 soft alignment sweep.}
    Values are measured after 1,000 matched joint-QAT steps on 64 held-out
    clips. Moment KL and mean gap are lower-is-better; an SD ratio of one
    matches the pristine FP32 latent spread.}
    \label{tab:appendix-lambda-sweep}
    \scriptsize
    \setlength{\tabcolsep}{3.7pt}
    \renewcommand{\arraystretch}{1.06}
    \begin{tabular*}{\textwidth}{@{\extracolsep{\fill}}lrrrrr@{}}
        \toprule
        $\lambda$ & Val. PSNR (dB) & SD ratio & Max SD & Moment KL
            & Mean gap \\
        \midrule
        0    & 29.906 & 1.922 & 2.133 & 0.9296 & 0.3973 \\
        0.03 & 30.103 & 1.088 & 1.121 & 0.0088 & 0.0238 \\
        0.1  & 29.969 & 1.031 & 1.052 & 0.0019 & 0.0235 \\
        0.3  & 29.768 & 1.009 & 1.022 & 0.0009 & 0.0227 \\
        1    & 29.500 & 0.999 & 1.011 & 0.0008 & 0.0219 \\
        3    & 29.309 & 0.994 & 1.006 & 0.0009 & 0.0222 \\
        10   & 29.191 & 0.995 & 1.007 & 0.0009 & 0.0221 \\
        \bottomrule
    \end{tabular*}
\end{table}

On the matched paired replay, $\lambda=0$ gives $(d_z,d_a)=(1.115,0.071)$,
whereas $\lambda\in[0.03,10]$ gives $d_z\in[0.158,0.222]$ and
$d_a\in[0.023,0.040]$. At $\lambda=0.03$, joint alignment improves both PSNR and interface preservation over $\lambda=0$, while larger weights reduce PSNR. These diagnostics do not establish whether a tuned joint objective can match LatentQuant in downstream control.

\begin{figure}[!htbp]
    \centering
    \includegraphics[width=\textwidth]{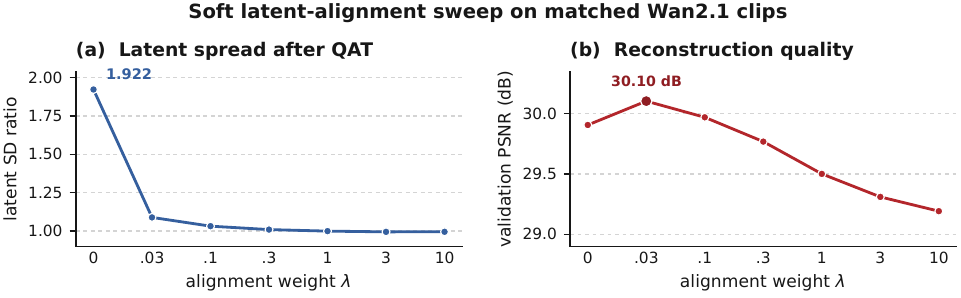}
    \caption{\textbf{Soft alignment sweep on matched Wan2.1 clips.}
    Increasing the alignment penalty reduces the latent standard-deviation
    ratio toward the FP32 reference, while held-out reconstruction PSNR
    decreases once the interface is tightly constrained.}
    \label{fig:lambda-sweep}
\end{figure}

\subsection{Reconstruction and Interface Metrics}
\label{app:evaluation-metrics}

Reconstruction is evaluated on fixed lists of 128 held-out clips from the
robot-manipulation distribution used for QAT and 128 clips reconstructed from
LIBERO rollouts. Following the component definitions of
VBench~\citep{huang2024vbench}, VBench-7 is the unweighted mean of subject consistency,
background consistency, temporal flickering, motion smoothness, dynamic
degree, aesthetic quality, and imaging quality. We use it as a compact summary
of reconstructed-video quality rather than as an official VBench composite
score.

For paired interface evaluation, all methods receive the same observation,
task condition, action noise, and frozen-policy state. Let $z_0$ and $a_0$
denote the high-precision latent and standardized action output, and let $z_q$
and $a_q$ denote their low-bit counterparts. The main paper reports the
channel-normalized latent RMSE $d_z$ and standardized action RMSE $d_a$.
Appendix~\ref{app:interface-replay} additionally reports ordinary relative
$\ell_2$ differences. The paired replay uses 20 latent requests and 10 action
requests for Wan2.1, and five latent requests and five action requests for
Wan2.2. These replay metrics isolate interface compatibility before
closed-loop trajectories diverge.

\subsection{Downstream Evaluation Protocols}
\label{app:downstream-protocols}

LIBERO uses 20 trials for each of ten tasks in every suite, giving 200 trials
per suite and 800 trials per method. RoboTwin is evaluated on 50 tasks with
10 rollouts per task, for 500 rollouts per method. All compared VAE conditions
use the same task lists, reset logic, action model, and success evaluator.

The physical evaluation uses the same frozen action model, camera calibration,
action horizon, control frequency, reset procedure, and instruction set for
every VAE condition. Fixed-observation replay is evaluated separately from
interactive real-robot evaluation so that latent and action deviations are not
confounded by diverging trajectories. Each physical task is decomposed into
ordered manipulation milestones with a fixed point value. Real-robot progress
is the sum of milestone points reached before timeout.

\section{Complete Reconstruction Results}
\label{app:vbench-full}

Table~\ref{tab:appendix-vbench-full} reports every component of the VBench-7
mean for the high-precision, Direct NVFP4, and LatentQuant deployment settings.
All methods within a VAE and video source are evaluated on the same 128 clips.
The abbreviations are SC (subject
consistency), BC (background consistency), TF (temporal flickering), MS
(motion smoothness), DD (dynamic degree), AQ (aesthetic quality), and IQ
(imaging quality). The table is provided to expose the complete metric profile;
the main claims do not require every individual dimension to improve.

\begin{table}[!htbp]
    \centering
    \caption{\textbf{Complete seven-dimension VBench reconstruction results.}
    The mean is unweighted, and each setting contains 128 reconstructed clips.}
    \label{tab:appendix-vbench-full}
    \scriptsize
    \setlength{\tabcolsep}{3.2pt}
    \renewcommand{\arraystretch}{1.04}
    \begin{tabular*}{\textwidth}{@{\extracolsep{\fill}}lllrrrrrrrr@{}}
        \toprule
        VAE & Video source & Method & SC & BC & TF & MS & DD & AQ & IQ & Mean \\
        \midrule
        Wan2.1 & Robot & \textsc{HP}
            & .9589 & .9732 & .9802 & .9906 & .3281 & .3514 & .6040 & .7409 \\
        Wan2.1 & Robot & \textsc{Direct}
            & .9408 & .9623 & .9750 & .9864 & .3203 & .3255 & .5760 & .7266 \\
        Wan2.1 & Robot & \textsc{Ours}
            & .9517 & .9673 & .9795 & .9901 & .3359 & .3508 & .5625 & .7340 \\
        Wan2.1 & LIBERO & \textsc{HP}
            & .9433 & .9638 & .9796 & .9931 & .5000 & .4246 & .6183 & .7747 \\
        Wan2.1 & LIBERO & \textsc{Direct}
            & .9319 & .9476 & .9755 & .9889 & .4531 & .3791 & .5830 & .7513 \\
        Wan2.1 & LIBERO & \textsc{Ours}
            & .9367 & .9548 & .9792 & .9918 & .4609 & .3723 & .5634 & .7513 \\
        \addlinespace[2pt]
        Wan2.2 & Robot & \textsc{HP}
            & .9593 & .9720 & .9804 & .9907 & .3359 & .3558 & .6110 & .7436 \\
        Wan2.2 & Robot & \textsc{Direct}
            & .9479 & .9622 & .9782 & .9887 & .3281 & .3365 & .5861 & .7325 \\
        Wan2.2 & Robot & \textsc{Ours}
            & .9587 & .9716 & .9808 & .9910 & .3359 & .3642 & .5872 & .7414 \\
        Wan2.2 & LIBERO & \textsc{HP}
            & .9446 & .9645 & .9798 & .9937 & .4922 & .4267 & .6236 & .7750 \\
        Wan2.2 & LIBERO & \textsc{Direct}
            & .9392 & .9510 & .9787 & .9916 & .4844 & .3930 & .6005 & .7626 \\
        Wan2.2 & LIBERO & \textsc{Ours}
            & .9453 & .9629 & .9803 & .9938 & .4766 & .4120 & .5885 & .7656 \\
        \bottomrule
    \end{tabular*}
\end{table}

\section{Additional Interface and Downstream Results}
\label{app:downstream-details}

\subsection{Matched Interface Replay}
\label{app:interface-replay}

Table~\ref{tab:appendix-interface-replay} expands the interface metrics used
in the main paper. Across both VAE families, \textsc{Joint} produces the
largest latent displacement and a correspondingly large change in the frozen
action output. \textsc{Ours} gives the smallest low-bit $d_z$ and $d_a$ in
both systems, including lower deviations than \textsc{Direct}; ordinary
relative $\ell_2$ differences show the same ordering.

\begin{table}[!htbp]
    \centering
    \caption{\textbf{Matched interface replay.}
    Relative $\ell_2$ differences are reported in addition to the normalized
    deviations used in the main paper. Lower is better.}
    \label{tab:appendix-interface-replay}
    \small
    \setlength{\tabcolsep}{6pt}
    \renewcommand{\arraystretch}{1.06}
    \begin{tabular*}{\textwidth}{@{\extracolsep{\fill}}llrrrr@{}}
        \toprule
        VAE & Method & $d_z$ & Latent rel. $\ell_2$
            & $d_a$ & Action rel. $\ell_2$ \\
        \midrule
        Wan2.1 & \textsc{HP}     & 0.000 & 0.000 & 0.000 & 0.000 \\
        Wan2.1 & \textsc{Direct} & 0.288 & 0.189 & 0.033 & 0.068 \\
        Wan2.1 & \textsc{Joint}  & 4.796 & 3.223 & 0.270 & 0.548 \\
        Wan2.1 & \textsc{J+KL}   & 1.190 & 0.803 & 0.118 & 0.241 \\
        Wan2.1 & \textsc{Ours}   & 0.202 & 0.135 & 0.018 & 0.036 \\
        \addlinespace[2pt]
        Wan2.2 & \textsc{HP}     & 0.000 & 0.000 & 0.000 & 0.000 \\
        Wan2.2 & \textsc{Direct} & 0.092 & 0.127 & 0.202 & 0.357 \\
        Wan2.2 & \textsc{Joint}  & 4.638 & 6.354 & 0.344 & 0.607 \\
        Wan2.2 & \textsc{J+KL}   & 0.995 & 1.360 & 0.340 & 0.600 \\
        Wan2.2 & \textsc{Ours}   & 0.062 & 0.086 & 0.003 & 0.005 \\
        \bottomrule
    \end{tabular*}
\end{table}

\subsection{LIBERO Suite Breakdown}
\label{app:libero-breakdown}

Table~\ref{tab:appendix-libero-breakdown} reports exact Wan2.1 success counts
for each LIBERO suite. The joint-QAT failure is not confined to one suite,
whereas \textsc{Ours} remains near the high-precision reference across all
four.

\begin{table}[!htbp]
    \centering
    \caption{\textbf{Wan2.1 LIBERO success by suite.}}
    \label{tab:appendix-libero-breakdown}
    \small
    \setlength{\tabcolsep}{4.3pt}
    \begin{tabular}{lrrrrr}
        \toprule
        Method & Spatial & Object & Goal & LIBERO-10 & Total \\
        \midrule
        \textsc{HP}     & 188/200 & 195/200 & 195/200 & 186/200 & 764/800 \\
        \textsc{Direct} & 187/200 & 195/200 & 190/200 & 183/200 & 755/800 \\
        \textsc{Joint}  &   3/200 &  65/200 &   0/200 &  16/200 &  84/800 \\
        \textsc{J+KL}   & 177/200 & 180/200 & 160/200 & 177/200 & 694/800 \\
        \textsc{Ours}   & 190/200 & 194/200 & 191/200 & 191/200 & 766/800 \\
        \bottomrule
    \end{tabular}
\end{table}

\subsection{RoboTwin Task Breakdown}
\label{app:robotwin-full}

RoboTwin is evaluated on 50 tasks with 10 rollouts per task. The aggregate
results are $349/500$ ($69.8\%$) for \textsc{HP}, $340/500$ ($68.0\%$)
for \textsc{Direct}, and $344/500$ ($68.8\%$) for \textsc{Ours}.
Table~\ref{tab:appendix-robotwin-tasks} reports all task-level counts under
the same task list, episode budget, and success evaluator.

\begin{table}[!htbp]
    \centering
    \caption{\textbf{Wan2.2 RoboTwin success by task.}
    Each task uses 10 rollouts. The two halves report high precision
    (\textsc{HP}), Direct NVFP4 (\textsc{D}), and two-stage QAT
    (\textsc{O}).}
    \label{tab:appendix-robotwin-tasks}
    \tiny
    \setlength{\tabcolsep}{2.4pt}
    \renewcommand{\arraystretch}{1.04}
    \resizebox{\textwidth}{!}{%
    \begin{tabular}{lrrr@{\hspace{10pt}}lrrr}
        \toprule
        Task & HP & D & O & Task & HP & D & O \\
        \midrule
        adjust\_bottle & 9/10 & 10/10 & 9/10
            & place\_can\_basket & 4/10 & 4/10 & 3/10 \\
        beat\_block\_hammer & 9/10 & 9/10 & 10/10
            & place\_cans\_plasticbox & 10/10 & 10/10 & 10/10 \\
        blocks\_ranking\_rgb & 5/10 & 2/10 & 2/10
            & place\_container\_plate & 9/10 & 9/10 & 9/10 \\
        blocks\_ranking\_size & 1/10 & 1/10 & 1/10
            & place\_dual\_shoes & 7/10 & 7/10 & 3/10 \\
        click\_alarmclock & 8/10 & 8/10 & 8/10
            & place\_empty\_cup & 9/10 & 9/10 & 8/10 \\
        click\_bell & 10/10 & 10/10 & 10/10
            & place\_fan & 9/10 & 9/10 & 10/10 \\
        dump\_bin\_bigbin & 10/10 & 7/10 & 9/10
            & place\_mouse\_pad & 8/10 & 8/10 & 8/10 \\
        grab\_roller & 10/10 & 10/10 & 10/10
            & place\_object\_basket & 5/10 & 4/10 & 3/10 \\
        handover\_block & 10/10 & 10/10 & 9/10
            & place\_object\_scale & 5/10 & 6/10 & 5/10 \\
        handover\_mic & 6/10 & 4/10 & 4/10
            & place\_object\_stand & 4/10 & 8/10 & 6/10 \\
        hanging\_mug & 3/10 & 4/10 & 4/10
            & place\_phone\_stand & 6/10 & 5/10 & 6/10 \\
        lift\_pot & 4/10 & 5/10 & 4/10
            & place\_shoe & 10/10 & 10/10 & 10/10 \\
        move\_can\_pot & 8/10 & 9/10 & 9/10
            & press\_stapler & 9/10 & 8/10 & 8/10 \\
        move\_pillbottle\_pad & 9/10 & 10/10 & 10/10
            & put\_bottles\_dustbin & 5/10 & 0/10 & 0/10 \\
        move\_playingcard\_away & 10/10 & 10/10 & 10/10
            & put\_object\_cabinet & 0/10 & 1/10 & 1/10 \\
        move\_stapler\_pad & 5/10 & 7/10 & 8/10
            & rotate\_qrcode & 8/10 & 7/10 & 9/10 \\
        open\_laptop & 9/10 & 9/10 & 9/10
            & scan\_object & 0/10 & 0/10 & 0/10 \\
        open\_microwave & 4/10 & 6/10 & 4/10
            & shake\_bottle & 10/10 & 9/10 & 10/10 \\
        pick\_diverse\_bottles & 6/10 & 5/10 & 7/10
            & shake\_bottle\_horizontally & 9/10 & 10/10 & 9/10 \\
        pick\_dual\_bottles & 7/10 & 8/10 & 9/10
            & stack\_blocks\_three & 9/10 & 9/10 & 10/10 \\
        place\_a2b\_left & 3/10 & 4/10 & 3/10
            & stack\_blocks\_two & 9/10 & 8/10 & 9/10 \\
        place\_a2b\_right & 2/10 & 1/10 & 3/10
            & stack\_bowls\_three & 9/10 & 6/10 & 6/10 \\
        place\_bread\_basket & 7/10 & 10/10 & 7/10
            & stack\_bowls\_two & 8/10 & 8/10 & 10/10 \\
        place\_bread\_skillet & 9/10 & 7/10 & 10/10
            & stamp\_seal & 9/10 & 8/10 & 9/10 \\
        place\_burger\_fries & 9/10 & 8/10 & 9/10
            & turn\_switch & 5/10 & 3/10 & 4/10 \\
        \midrule
        \multicolumn{4}{r}{Aggregate: \textsc{HP} $349/500$}
            & \multicolumn{4}{l}{\textsc{Direct} $340/500$;
            \textsc{Ours} $344/500$} \\
        \bottomrule
    \end{tabular}}
\end{table}

\subsection{Open-Loop Action Traces}
\label{app:openloop-action}

\begin{figure}[!htbp]
    \centering
    \includegraphics[width=\linewidth]{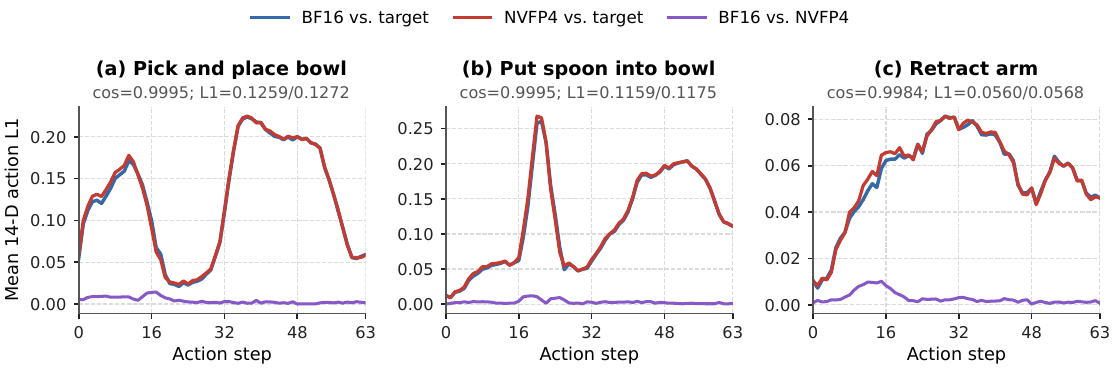}
    \caption{\textbf{Representative Wan2.2 open-loop action traces.}
    Blue and red report mean absolute action error against the same reference
    trajectory for the BF16 and two-stage NVFP4 VAEs; purple reports their
    direct prediction difference.}
    \label{fig:appendix-openloop-action}
\end{figure}

Figure~\ref{fig:appendix-openloop-action} shows three representative
64-step action traces under a fixed Wan2.2 action model. The high-precision
and two-stage NVFP4 VAEs receive identical observations, prompts, noise, and
policy parameters. Across the complete five-event paired set, the
high-precision and NVFP4 action L1 errors are $0.121864$ and $0.121618$,
respectively, and the mean prediction cosine is $0.995265$. This diagnostic
measures paired action compatibility and is not a substitute for closed-loop
success.

\subsection{Physical-Evaluation Breakdown}
\label{app:real-robot-protocol}

The physical evaluation follows the protocol specified in
Appendix~\ref{app:downstream-protocols}. Fixed-observation replay and
interactive real-robot evaluation are reported separately. Each task is scored
with the same predefined milestone-point rubric. Each method is evaluated in
three complete rounds over the same 15 tasks, for 45 physical task executions
per method. Table~\ref{tab:appendix-real-robot-scores} reports all selected
round-level scores. Across the 15 task-wise three-round means, LatentQuant is lower on
7 tasks, higher on 6, and unchanged on 2. The median paired difference
is $0.00$ points, the mean paired difference is $+0.43$ points, and the
mean absolute difference is $2.02$ points. We therefore interpret these
results as parity rather than an improvement from low-precision execution.

\begin{table}[t]
\centering
\caption{
\textbf{Complete real-robot evaluation breakdown.}
Each task is evaluated over three independent rounds under the same
physical-robot protocol. Each entry is the accumulated task score
(maximum 100). ``Mean'' averages the three rounds, and
$\Delta$ denotes LatentQuant minus FP32.
}
\label{tab:appendix-real-robot-scores}
\scriptsize
\setlength{\tabcolsep}{2.6pt}
\renewcommand{\arraystretch}{1.04}
\resizebox{\textwidth}{!}{%
\begin{tabular}{lrrrrrrrrr}
\toprule
&
\multicolumn{4}{c}{FP32}
&
\multicolumn{4}{c}{LatentQuant}
& \\
\cmidrule(lr){2-5}
\cmidrule(lr){6-9}
Task
& R1 & R2 & R3 & Mean
& R1 & R2 & R3 & Mean
& $\Delta$ \\
\midrule

put\_spoon\_to\_bowl
& 2.00 & 2.00 & 0.00 & 1.33
& 0.00 & 0.00 & 2.00 & 0.67
& -0.67 \\

put\_glasses\_on\_woodshelf
& 0.00 & 6.00 & 4.00 & 3.33
& 0.00 & 0.00 & 0.00 & 0.00
& -3.33 \\

put\_items\_into\_drawer
& 0.00 & 0.00 & 0.00 & 0.00
& 0.00 & 0.00 & 0.00 & 0.00
& 0.00 \\

pick\_items\_into\_basket
& 19.25 & 12.75 & 21.75 & 17.92
& 22.00 & 8.50 & 13.50 & 14.67
& -3.25 \\

pick\_fruits\_into\_basket
& 13.00 & 15.00 & 11.00 & 13.00
& 11.00 & 32.00 & 31.00 & 24.67
& +11.67 \\

pour\_water\_from\_bottle
& 4.00 & 0.00 & 0.00 & 1.33
& 0.00 & 0.00 & 0.00 & 0.00
& -1.33 \\

sort\_headphone
& 33.00 & 25.00 & 33.00 & 30.33
& 33.00 & 38.00 & 24.00 & 31.67
& +1.33 \\

arrange\_cup\_inverted\_triangle
& 3.00 & 1.00 & 3.00 & 2.33
& 2.00 & 0.00 & 4.00 & 2.00
& -0.33 \\

insert\_wireline
& 3.00 & 0.00 & 3.00 & 2.00
& 0.00 & 0.00 & 3.00 & 1.00
& -1.00 \\

put\_ring\_onto\_rod
& 6.00 & 0.00 & 0.00 & 2.00
& 3.00 & 6.00 & 0.00 & 3.00
& +1.00 \\

press\_button\_in\_order
& 0.00 & 0.00 & 6.00 & 2.00
& 0.00 & 0.00 & 0.00 & 0.00
& -2.00 \\

put\_stationery\_in\_case
& 0.00 & 2.00 & 2.00 & 1.33
& 0.00 & 5.00 & 3.00 & 2.67
& +1.33 \\

classify\_items\_as\_shape
& 6.00 & 15.00 & 3.00 & 8.00
& 6.00 & 12.00 & 6.00 & 8.00
& 0.00 \\

put\_blocks\_to\_color
& 3.00 & 0.00 & 0.00 & 1.00
& 6.00 & 0.00 & 0.00 & 2.00
& +1.00 \\

pair\_up\_items
& 0.00 & 12.00 & 4.00 & 5.33
& 6.00 & 6.00 & 10.00 & 7.33
& +2.00 \\

\midrule
\multicolumn{10}{l}{\textit{Task-wise summary:}
7 lower, 2 tied, and 6 higher; median $\Delta=0.00$;
mean $\Delta=+0.43$.} \\

\bottomrule
\end{tabular}}
\end{table}

\section{Additional Mechanism Analysis}
\label{app:mechanism-details}

The supplementary mechanism evidence is ordered by inferential strength.
Appendix~\ref{app:layerwise-transport} reports controlled layerwise
interventions; Appendix~\ref{app:reverse-vjp} isolates the one-site causal
chain and verifies reference invariance of the total gradient change.
The remaining diagnostics show matched training trajectories and their
distribution across time and channels, but are not used as independent causal
estimators.

\subsection{Layerwise Decoder Transport}
\label{app:layerwise-transport}

For the 38-site decoder sweep, QDQ sites are ordered from the reconstruction
output toward the latent input. Starting from the FP32 decoder, we replace one
additional activation site at a time and record reconstruction, loss, latent
gradient, and latent-RMS pressure. The difference between adjacent hybrid
branches is the marginal contribution assigned to the newly enabled site.
Summing all signed increments recovers the measured difference between the
full-FP32 and full-QDQ endpoints. This telescoping construction preserves
interactions with sites already replaced and should not be interpreted as a
collection of independent one-site effects.

The forward sweep shows that local QDQ error remains near $9.2\%$, while
decoder residual and nonlinear paths can amplify or cancel previously
introduced perturbations. Full decoder QDQ yields an $11.04\%$ reconstruction
difference and changes the loss from $0.008834$ to $0.017591$. The backward
sweep shows that the corresponding output gradient has cosine $0.158$ with its
FP32 counterpart and the latent gradient has cosine $0.296$; their norm ratios
are $1.715\times$ and $1.375\times$, respectively. The nonmonotonic
layerwise curves are expected because newly introduced vector perturbations can
reinforce or cancel the accumulated differential signal.

\begin{figure}[!htbp]
    \centering
    \includegraphics[width=\textwidth]{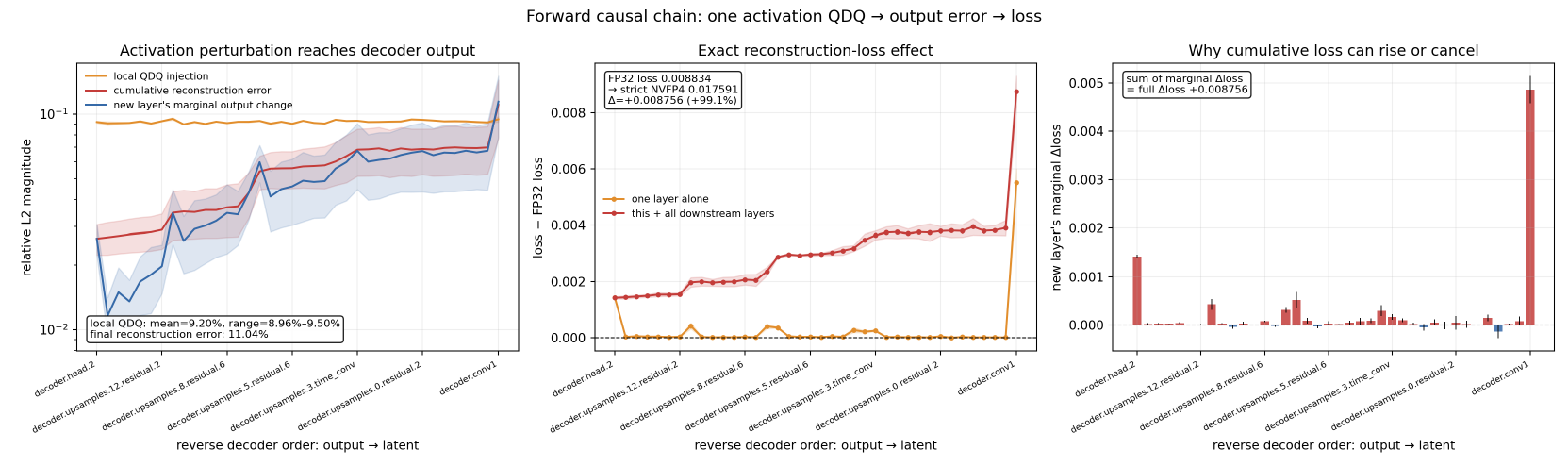}

    \vspace{0.6em}

    \includegraphics[width=\textwidth]{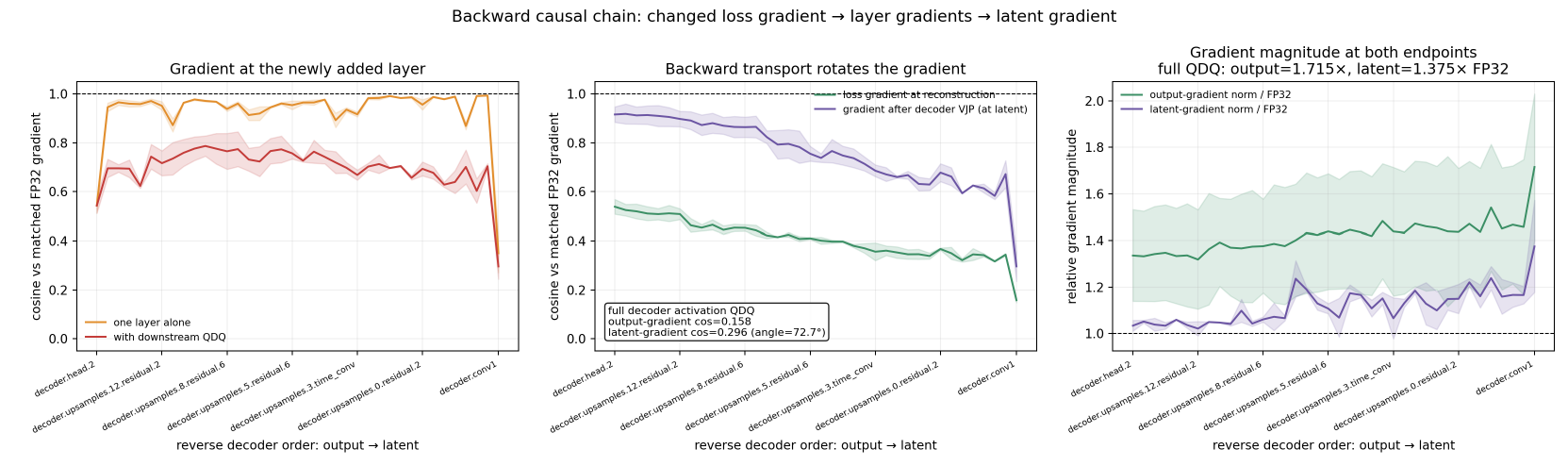}
    \caption{\textbf{Layerwise forward and backward transport.}
    Top: local QDQ injection, accumulated reconstruction difference, isolated
    and cumulative loss changes, and signed telescoping increments.
    Bottom: isolated and cumulative gradient changes and their transport to the
    latent through the decoder VJP. Shaded regions show variation across the
    paired clips.}
    \label{fig:appendix-layerwise-transport}
    \label{fig:appendix-layerwise-forward}
    \label{fig:appendix-layerwise-backward}
\end{figure}

\subsection{One-Site Transport and Reverse-Reference Decomposition}
\label{app:reverse-vjp}

Table~\ref{tab:appendix-one-site-chain} reports the intervention used for the
main causal trace. Only the activation entering the first decoder convolution
is quantized; all weights and encoder activations remain FP32, and the decoder
is frozen. The table deliberately separates this matched one-site chain from
the full-decoder RMS-pressure endpoint reported in the main text.

\begin{table}[!htbp]
    \centering
    \caption{\textbf{Measured one-site causal chain.}
    Gradient angles and norm ratios are relative to the matched FP32 branch.}
    \label{tab:appendix-one-site-chain}
    \small
    \setlength{\tabcolsep}{5pt}
    \begin{tabular}{lr}
        \toprule
        Quantity & Measurement \\
        \midrule
        Local activation perturbation & $9.708\%$ \\
        First-convolution output difference & $8.315\%$ \\
        Final reconstruction difference & $11.857\%$ \\
        Loss increase & $0.006120$ \\
        Output-gradient cosine / norm ratio & $0.3155$ / $1.640\times$ \\
        Latent-gradient angle / norm ratio & $67.45^\circ$ / $1.126\times$ \\
        Encoder-gradient angle / norm ratio & $32.49^\circ$ / $1.514\times$ \\
        VJP decomposition closure error & $0.067\%$ \\
        \bottomrule
    \end{tabular}
\end{table}

Equation~\ref{eq:gradient_decomposition} uses the FP32 decoder as the reference
backward map. The same latent-gradient difference admits the reverse-reference
form
\begin{equation}
    g_z^Q-g_z^F
    =
    J_Q^\top(u_Q-u_F)
    +
    (J_Q^\top-J_F^\top)u_F .
    \label{eq:reverse-vjp}
\end{equation}
The allocation between loss-gradient transport and backward-map shift changes
with the reference choice, but their sum, the closure error, and the measured
rotation of $g_z$ remain unchanged. We therefore base the causal claim on the total differential signal, using the decomposition to distinguish reconstruction-signal and backward-map changes rather than quantify their relative contributions.

\subsection{Matched Training Trajectories}
\label{app:matched-trajectories}

\begin{figure}[!htbp]
    \centering
    \includegraphics[width=0.98\textwidth]{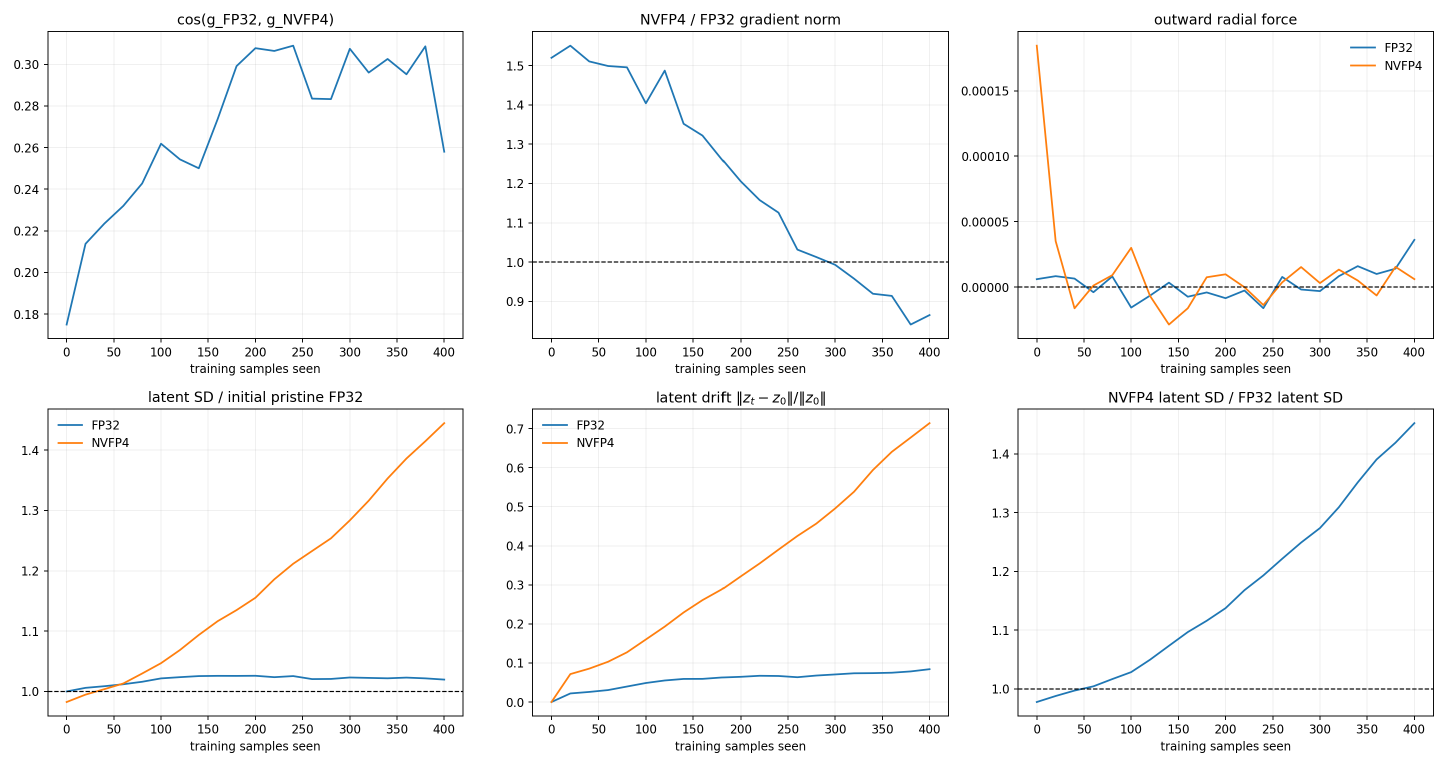}
    \caption{\textbf{Matched FP32 and strict-NVFP4 training trajectories.}
    The top row reports gradient direction, norm ratio, and the instantaneous
    radial-force proxy. The bottom row reports latent standard deviation,
    sample-wise latent drift, and their cross-condition ratio.}
    \label{fig:appendix-training-trajectory}
\end{figure}

\paragraph{Global trajectories.}
Figure~\ref{fig:appendix-training-trajectory} reports a longer matched
diagnostic in which FP32 and strict-NVFP4 branches share initialization,
samples, loss, and optimizer. Latent standard deviation and sample-wise drift
diverge persistently even when the instantaneous radial-force proxy fluctuates
around zero. This is supplementary trajectory evidence; causal localization is
provided by the decoder-only intervention above.

\begin{figure}[!htbp]
    \centering
    \includegraphics[width=0.72\linewidth]{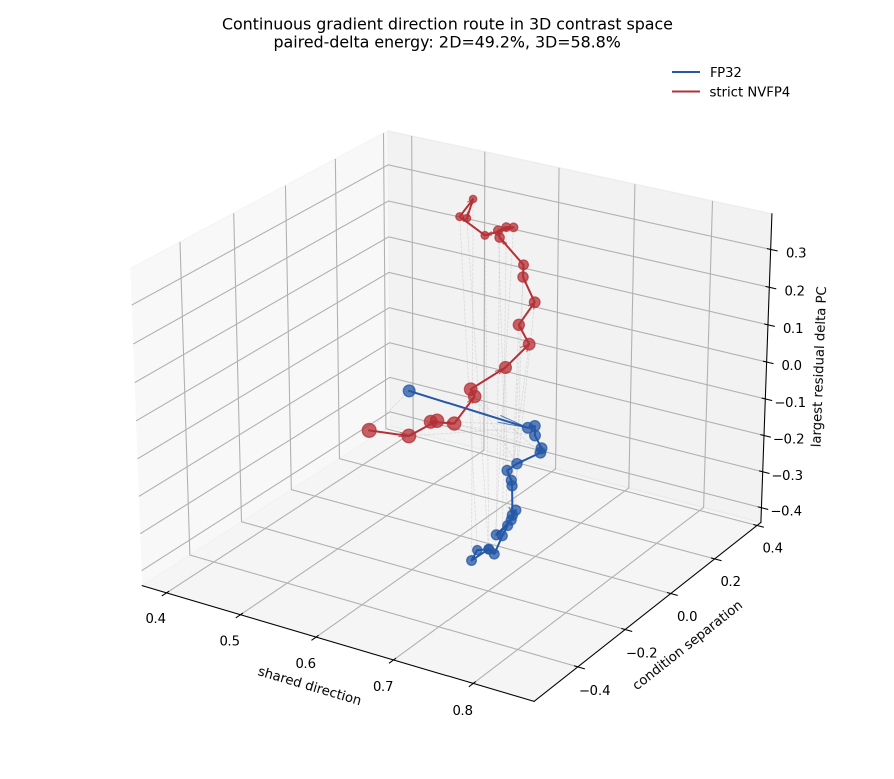}
    \caption{\textbf{Directional visualization of matched gradients.}
    Connected markers follow training order for the FP32 and strict-NVFP4
    gradient-direction trajectories in the three-dimensional contrast basis.}
    \label{fig:appendix-gradient-pca}
\end{figure}

\paragraph{Directional projection.}
Figure~\ref{fig:appendix-gradient-pca} projects the same paired trajectories
into a three-dimensional contrast basis. The third axis is selected from the
residual paired difference after defining a shared direction and the leading
condition-separation direction. The projection captures $58.8\%$ of
paired-difference energy. It is included as a visualization of persistent
directional separation and is not used to estimate a causal effect.

\begin{figure}[!htbp]
    \centering
    \includegraphics[width=0.94\textwidth]{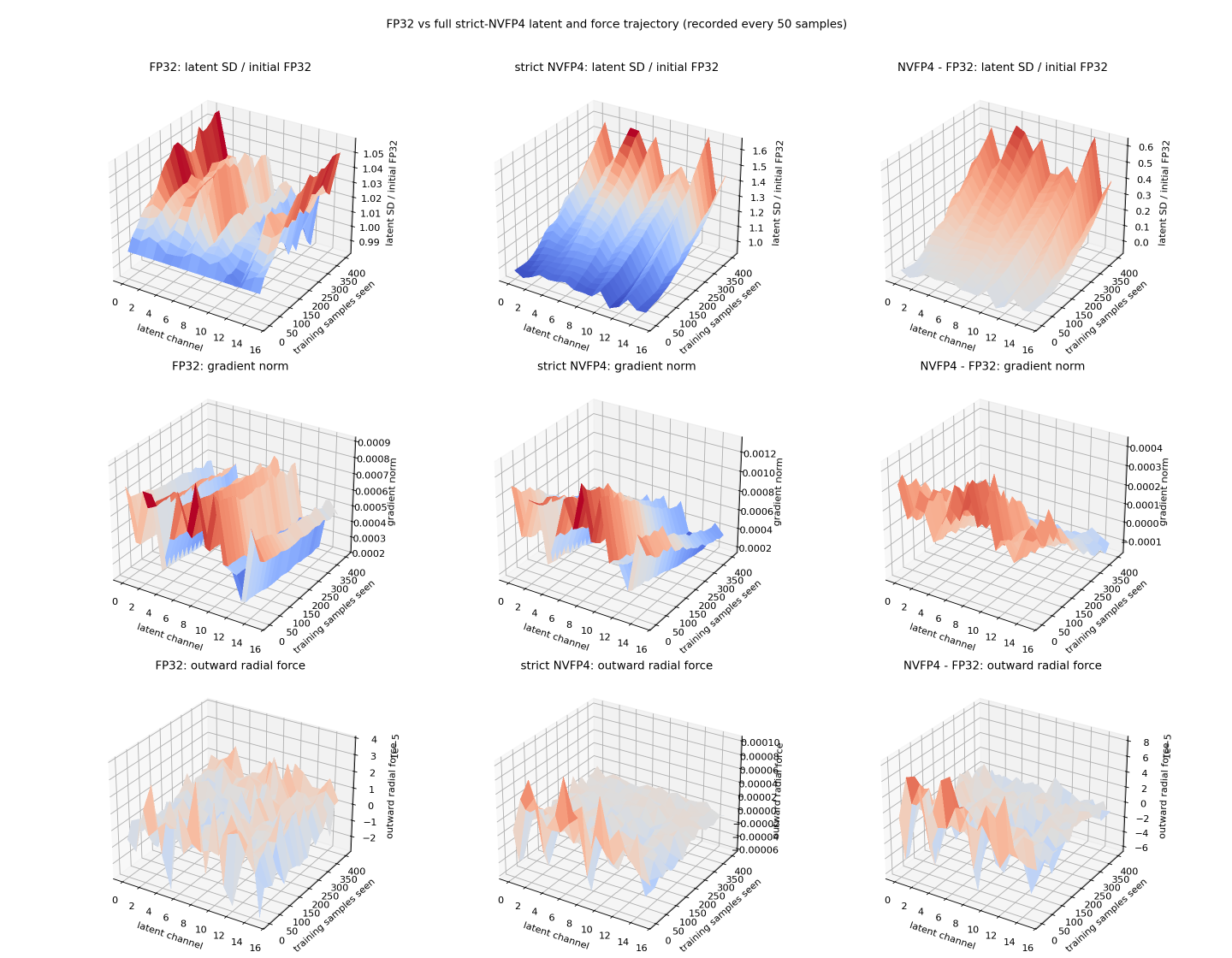}
    \caption{\textbf{Channel-wise strict-NVFP4 minus FP32 training
    trajectories.}
    The panels report latent standard-deviation ratio, latent-gradient norm,
    and RMS-directed-force differential for all 16 latent channels, measured
    every 20 training samples.}
    \label{fig:appendix-channel-trajectory}
\end{figure}

\paragraph{Channel-wise view.}
Figure~\ref{fig:appendix-channel-trajectory} resolves the matched diagnostic
by latent channel and training time. Each panel isolates the strict-NVFP4
minus FP32 differential rather than repeating the global trajectories.
Latent growth is distributed across channels rather than confined to a single
projection coordinate. The gradient-norm and RMS-directed-force maps reveal
temporal heterogeneity and should not be interpreted as independent
per-channel causal effects.

\subsection{Temporal-Length Control}
\label{app:temporal-generalization}

The QAT clips contain 17 frames, whereas deployment may use longer sequences.
We therefore repeat paired reconstruction checks at 33 and 65 frames without
retraining. From 17 to 65 frames, the NVFP4--high-precision PSNR gap widens by
only $0.273$ dB for Wan2.1 and $0.108$ dB for Wan2.2, and the paired VBench
diagnostic shows no systematic widening. This control evaluates reconstruction across input lengths, not error accumulation over sequential task chains.

\section{Native NVFP4 Backend and Measurement Protocol}
\label{app:systems-details}

\subsection{End-to-End Timing}
\label{app:end-to-end-timing}

The native backend stores supported weights in packed form and dispatches
operator-specialized convolution kernels. Unsupported shapes retain an
explicit higher-precision fallback. Timing uses synchronized warm-up and
repeated measurements at batch size~1. Weights are packed offline; activation
scaling and packing remain inside the measured NVFP4 path.
The headline 17-frame throughput gains use dedicated clean-device repeats at
the same 1800~MHz clock: $56.499$ versus $47.629$~ms for Wan2.1 and $65.755$
versus $51.499$~ms for Wan2.2. Table~\ref{tab:appendix-temporal-latency}
reports a separate temporal-length sweep used to assess shape scaling rather
than to recompute those headline percentages; this sweep supplies the values
shown in Figure~\ref{fig:nvfp4_efficiency}.

\begin{table}[!htbp]
    \centering
    \caption{\textbf{End-to-end VAE encode--decode latency on NVIDIA B300.}
    Resolution is $208\times224$, batch size is one, and values are medians in
    milliseconds.}
    \label{tab:appendix-temporal-latency}
    \small
    \setlength{\tabcolsep}{5.2pt}
    \renewcommand{\arraystretch}{1.06}
    \begin{tabular*}{\textwidth}{@{\extracolsep{\fill}}lrrrrrrr@{}}
        \toprule
        VAE & Frames & BF16 enc. & BF16 dec. & NVFP4 enc. & NVFP4 dec.
            & Total reduction & Speedup \\
        \midrule
        Wan2.1 & 17 & 22.970 & 34.079 & 19.954 & 27.978 & $16.08\%$ & $1.192\times$ \\
        Wan2.1 & 33 & 40.854 & 60.905 & 36.104 & 50.242 & $15.16\%$ & $1.179\times$ \\
        Wan2.1 & 65 & 76.608 & 113.663 & 68.378 & 94.420 & $14.45\%$ & $1.169\times$ \\
        Wan2.2 & 17 & 26.316 & 41.702 & 23.486 & 31.524 & $19.13\%$ & $1.237\times$ \\
        Wan2.2 & 33 & 43.522 & 75.752 & 38.778 & 56.129 & $20.43\%$ & $1.257\times$ \\
        Wan2.2 & 65 & 91.246 & 146.416 & 81.915 & 108.366 & $19.97\%$ & $1.249\times$ \\
        \bottomrule
    \end{tabular*}
\end{table}

Wan2.1 uses 20 warm-up and 100 timed iterations; Wan2.2 uses 10 warm-up and
50 timed iterations. The GPU clock is fixed at 1800 MHz. Each measurement
includes the complete VAE graph, including the higher-precision
normalization, nonlinear, residual, and boundary operations.

\subsection{Stage and Operator Breakdown}
\label{app:operator-breakdown}

\begin{figure}[!htbp]
    \centering
    \includegraphics[width=\textwidth]{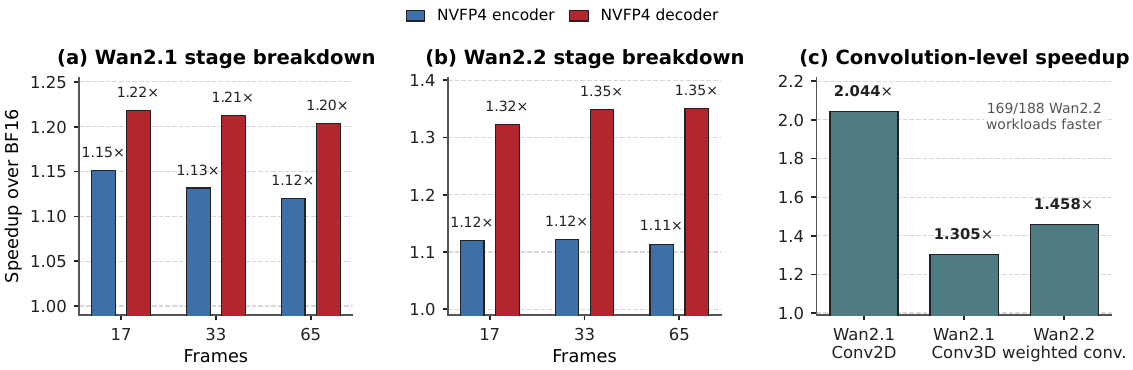}
    \caption{\textbf{Stage- and operator-level acceleration over BF16 cuDNN.}
    Panels (a) and (b) report encoder and decoder speedups from the complete
    temporal sweep. Panel (c) reports a $2.044\times$ Wan2.1 Conv2D-region
    speedup, a $1.305\times$ Wan2.1 Conv3D-region speedup, and a
    $1.458\times$ call-count-weighted Wan2.2 convolution speedup; 169 of 188
    measured Wan2.2 workloads are faster under NVFP4.}
    \label{fig:appendix-operator-breakdown}
\end{figure}

Figure~\ref{fig:appendix-operator-breakdown} resolves the end-to-end result
into encoder and decoder stages. Both stages are faster at every tested
temporal length, with the larger gain in the decoder. The convolution
diagnostics use their native profiling scopes: Wan2.1 aggregates complete
Conv2D and Conv3D regions, whereas the Wan2.2 value is call-count weighted
over 188 observed convolution workloads. Of these workloads, 169 are faster
under the native NVFP4 path. These diagnostics explain the end-to-end trend
but do not replace the graph-level measurements in
Table~\ref{tab:appendix-temporal-latency}.

\subsection{Resident State and DRAM Traffic}
\label{app:system-resources}

The memory quantities in Table~\ref{tab:appendix-system-resources} are
scope-qualified. Resident model state counts unique unquantized VAE state and
one packed payload per quantized convolution after releasable convolution
masters are removed; it excludes shape-plan activation and output scratch.
DRAM traffic is summed over all kernels in one profiler-delimited
encode--decode graph and denotes transferred volume rather than physical
bandwidth.

\begin{table}[!htbp]
    \centering
    \caption{\textbf{Measured deployment resources.}
    Batch size is one and temporal length is 17 frames. Each row compares BF16
    with the packed NVFP4 execution path.}
    \label{tab:appendix-system-resources}
    \small
    \setlength{\tabcolsep}{6pt}
    \begin{tabular*}{\textwidth}{@{\extracolsep{\fill}}llrrrc@{}}
        \toprule
        VAE & Metric & BF16 & NVFP4 & Reduction & Unit \\
        \midrule
        Wan2.1 & Resident VAE model state & 242.03 & 95.93 & $60.36\%$ & MiB \\
        Wan2.1 & Full-VAE DRAM traffic & 25.176 & 19.258 & $23.51\%$ & GiB \\
        Wan2.2 & Resident VAE model state & 1344.09 & 410.68 & $69.45\%$ & MiB \\
        Wan2.2 & Full-VAE DRAM traffic & 37.127 & 15.588 & $58.01\%$ & GiB \\
        \bottomrule
    \end{tabular*}
\end{table}

The ideal packed payload is 4.5 bits per value: four E2M1 data bits plus one
8-bit block scale per 16 values, excluding negligible tensor-scale
amortization. This format-level value differs from the measured resident-state
reduction because higher-precision boundaries and unsupported operators remain
in the deployed VAE.

\subsection{Correctness and Fallback Policy}
\label{app:backend-correctness}

Before performance measurement, CUDA QDQ is compared with the reference
implementation across operand ranks, channel placements, input dtypes, zero
tensors, and E2M1/E4M3 midpoint cases. End-to-end full-sequence and chunked
Wan2.2 execution are also compared at 17 frames; measured latent and
reconstruction differences remain within the configured numerical tolerances.
Native convolution dispatch is enabled only for validated shapes. Unsupported
workloads fall back explicitly rather than silently changing the numerical
contract.

\end{document}